\documentclass[10pt,twocolumn,letterpaper]{article} 

\usepackage{avss}
\usepackage{times}
\usepackage{epsfig}
\usepackage{graphicx}
\usepackage{amsmath}
\usepackage{amssymb}
\usepackage{adjustbox}
\usepackage[ruled,lined]{algorithm2e}
\usepackage{booktabs}       
\usepackage{amsfonts}       
\usepackage{nicefrac}       
\usepackage{microtype}      
\usepackage{graphicx}
\usepackage{makecell}
\usepackage{amsmath,amssymb,booktabs,tabulary,multirow,overpic,xcolor}
\usepackage{enumitem}
\usepackage{pifont}
\definecolor{dgreen}{rgb}{0.04,0.7,0.13}
\usepackage{float}
\newcommand{\cmark}{{\color{black}\ding{51}}}%
\newcommand{\xmark}{{\color{black}\ding{55}}}%

\usepackage{lipsum}
\usepackage[normalem]{ulem}

\usepackage[tight,footnotesize]{subfigure}

\usepackage{caption}

\usepackage[pagebackref=true,breaklinks=true,letterpaper=true,colorlinks,bookmarks=false]{hyperref}

\avssfinalcopy 
\def\avssPaperID{2} 

\makeatletter
\def\thanks#1{\protected@xdef\@thanks{\@thanks
		\protect\footnotetext{#1}}}

\makeatother

\ifavssfinal\fi

\begin{document}

	
	\title{Learning Temporal 3D Human Pose Estimation with Pseudo-Labels}
\author{\parbox{16cm}{\centering
		{\large Arij Bouazizi$^{1,2}$, Ulrich Kressel$^1$ and Vasileios Belagiannis$^2$}\\
		{\normalsize
			$^1$ Mercedes-Benz AG, Stuttgart, Germany \\
			$^2$ Universit\"at Ulm, Ulm, Germany.
	}}
	\thanks{E-mail: \textit{firstname.lastname@\{daimler.com, uni-ulm.de\}}.}
}

	\maketitle
	\thispagestyle{empty}
	
	\begin{abstract}
		We present a simple, yet effective, approach for self-supervised 3D human pose estimation. Unlike the prior work, we explore the temporal information next to the multi-view self-supervision. During training, we rely on triangulating 2D body pose estimates of a multiple-view camera system. A temporal convolutional neural network is trained with the generated 3D ground-truth and the geometric multi-view consistency loss, imposing geometrical constraints on the predicted 3D body skeleton. During inference, our model receives a sequence of 2D body pose estimates from a single-view to predict the 3D body pose for each of them. An extensive evaluation shows that our method achieves state-of-the-art performance in the Human3.6M and MPI-INF-3DHP benchmarks. Our code and models are publicly available at  \url{https://github.com/vru2020/TM_HPE/}. 

	\end{abstract}
	
	
	\section{Introduction}
	
	Estimating the 3D human body posture from an image is a long-standing problem in computer vision with many applications such as trajectory forecasting~\cite{hasan2019forecasting} and gesture recognition \cite{wiederer2020traffic}. The main research trend is the end-to-end 3D body pose estimation with deep neural networks~\cite{habibie2019wild,kanazawa2018end,kolotouros2019learning}. In this course, several approaches~\cite{pavlakos2017coarse,Martinez17,Pavllo19,tripathi2020posenet3d} adopt an off-the-shelf 2D body pose estimator to predict the 2D joint positions in the image space, followed by a 2D-3D lifting. Despite the fact that these approaches achieve promising results in standard benchmarks, most of them have the disadvantage of requiring ground-truth data. Acquiring 3D keypoints is not only expensive, but also hard to obtain due to the lack of the third dimension when annotating images. This bottleneck significantly hinders the application in unconstrained scenarios, since it requires annotating new data.
	
	Weakly and self-supervised learning methods relaxed the need of 3D ground-truth body poses  by exploiting unpaired 2D and 3D body poses~\cite{wandt2019repnet,drover20183d,tripathi2020posenet3d} or multi-view images \cite{bouazizi2021self,Kocabas19,kundu2020self}. Nevertheless, not a single method explores the temporal information next to the multi-view self-supervision. This work presents a simple and effective approach for temporal 3D human pose estimation using 3D pseudo-labels and a temporal model.
	
	We phrase the 3D human pose estimation problem as a 2D pose estimation followed by a 2D-3D lifting. During training, we rely on a multiple-view camera system and 2D body pose estimates from each camera view to create 3D pseudo-labels via triangulation. A temporal convolutional neural network \cite{Pavllo19} is then trained with the generated 3D ground-truth. To further constrain the 3D search space, we present the multi-view consistency objective as a  geometrical constraint on the predicted 3D body skeleton. During inference, our approach receives a sequence of 2D body pose estimates as input to predict the 3D body pose for each of them. It is important to note that a multiple-view configuration is only necessary during training.
	
	\begin{figure}
		\includegraphics[width=\linewidth]{{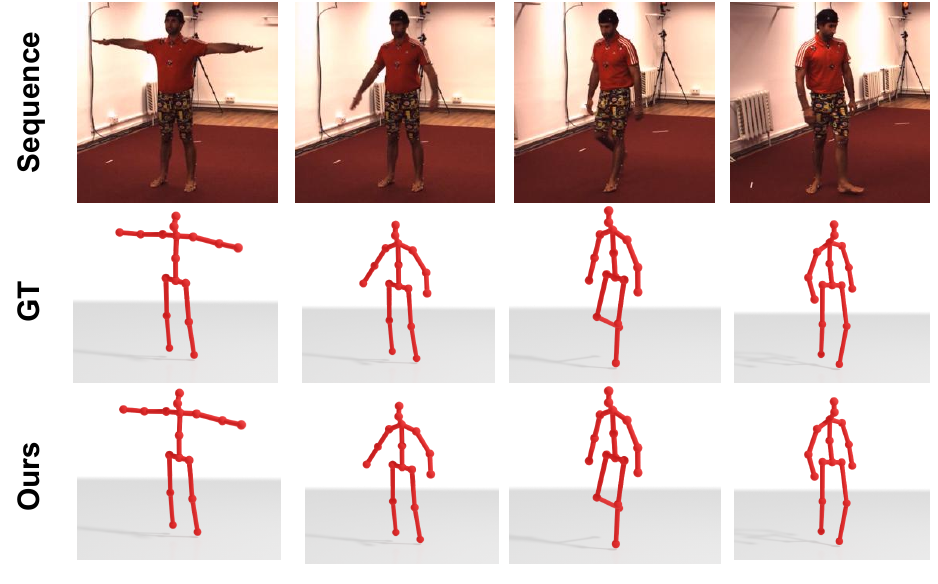}}
		\caption{Qualitative results on Human3.6M ~\cite{ionescu2013human3}. We present a self-supervised learning approach for single human 3D body pose estimation from sequences of 2D body pose estimates. Our model achieves competitive results with state-of-the art fully-supervised methods.}
	\end{figure}

	We empirically show the benefit of modeling the temporal information over single-frame approaches. We conduct an extensive evaluation on two publicly available benchmarks, Human3.6M~\cite{ionescu2013human3} and MPI-INF-3DHP~\cite{mpiiinfdataset}. Our approach achieves state-of-the-art performance on Human3.6M, improving upon the previous self-supervised approaches by $25.0\%$. Our results are also competitive to the fully supervised approaches, which rely on 3D ground-truth body poses and temporal information. On MPI-INF-3DHP, our method yields the lowest average error, which outperforms most state-of-the-art fully-supervised approaches  \cite{habibie2019wild,kanazawa2018end,kolotouros2019learning}.

	In the remainder of the paper, we first discuss the related 3D human pose estimation approaches. We then introduce our self-supervised video based approach and finally demonstrate that our model achieves competitive results compared with the state-of-the art fully-supervised methods on standard 3D human pose estimation benchmarks.
	

	\section{Related Work}
	\label{related_work}
	Similar to the literature \cite{wandt2021canonpose,kundu2020self,li2020geometry,bouazizi2021self}, the related work can be partitioned into supervised, unsupervised, self- and weakly-supervised learning. Below, we discuss the related approaches to our method.
	
	\paragraph{Supervised Learning}
	
	There is a vast literature on 3D human pose estimation based on ground-truth labels ~\cite{belagiannis2014holistic,zhou2016deep,Martinez17,Lu2018,pavlakos2017coarse,Pavllo19,zhou2016sparse,bogo2016smpl}. While the current state-of-the-art is completely based on deep neural networks ~\cite{zhou2016deep,pavlakos2017coarse,Pavllo19}, the prior work also includes graphical models with hand-crafted features, such as pictorial structures models \cite{Belagiannis2014}. Most of the currently top-performing methods build on top of an off-the-shelf 2D image-based pose estimator for 2D keypoint detection and then perform lifting to the 3D coordinate space~\cite{Martinez17,Pavllo19}. Also, there have been efforts in exploring temporal information with deep neural networks. For instance, Pavallo \textit{et.~al.} ~\cite{Pavllo19} show that video-based 3D body pose can be effectively estimated with a fully convolutional model based on dilated temporal convolutions over 2D keypoints. Despite the remarkable results, these approaches require ground-truth information during training. Our work reaches similar performance to supervised learning by only harnessing 3D pseudo-labels.
	
	\paragraph{Unsupervised Learning} 
	
	Unsupervised learning approaches omit the need of labeling the data. Adversarial learning, for instance compensates for the lack of ground-truth information~\cite{chenunsupervised19,tripathi2020posenet3d,kundu2020self}  by imposing a geometric prior on the 3D structure. Chen \textit{et.~al.}~\cite{chenunsupervised19} introduce the project-lift-project training strategy. The approach is motivated by the fact that predicted 3D skeletons can be randomly rotated and projected without any change in the distribution of the resulting 2D skeletons. While the adversarial training arguably removes the dependence on data annotation, it is inherently still ambiguous as multiple 3D poses can map to the same 2D keypoints. Wandt  \textit{et.~al.} ~\cite{wandt2021canonpose} propose to alleviate the modeling ambiguity of the 3D body pose by projecting the 2D detections from one view to another view via a canonical pose space. Noteworthy, these works, although effective, still need adversarial learning and are far inferior to fully supervised approaches. In this work, we present a simpler yet more effective approach, which learns a temporal model from body pose estimates and multiple-view geometry. As our experiments show, we achieve better performance without the need for adversarial learning.

	\begin{table}[t]
		\footnotesize
		\caption{  
			Characteristic comparison of our approach against prior weakly, self- and supervised approaches, in terms of different levels of supervision \vspace{0.8mm} }
		\centering
		\setlength\tabcolsep{3.0pt}
		\resizebox{0.47\textwidth}{!}{
			\begin{tabular}{l|cc|c|c}
				\hline
				\multirow{2}{*}{Methods} & \multicolumn{2}{c|}{ \makecell{Paired supervision\\(MV: muti-view)}} &
				\multirow{2}{*}{\makecell{\vspace{-3.5mm}\\Unpaired\\3D pose\\Supervision}} &
				\multirow{2}{*}{\makecell{\vspace{-3.5mm} \\ 3D pose \\ Ground-Truth \\ Supervision}} \\
				\cline{2-3}  
				& \makecell{MV\\ pair}  & \makecell{2D \\pose} & 
				\\ \hline\hline
				Pavllo \etal~\cite{Pavllo19} &\xmark  &\cmark &\xmark &\cmark    \\
				Martinez \etal~\cite{Martinez17}  &\xmark &\cmark &\xmark &\cmark    \\
				
				Kocabas \etal~\cite{Kocabas19} &\cmark  &\cmark &\xmark &\xmark     \\

				Wandt \etal~\cite{wandt2019repnet} &\xmark  &\cmark &\cmark &\xmark    \\
				Chen \etal~\cite{chenunsupervised19} &\xmark  &\cmark &\cmark &\xmark    \\
				Tripathi \etal~\cite{tripathi2020posenet3d} &\xmark  &\cmark &\cmark &\xmark    \\
				
				\hline
				Ours  &\cmark  &\cmark &\xmark &\xmark     \\
				\hline
		\end{tabular}}
		\vspace{-2mm}
		\label{tab:comparison}
	\end{table}

	\paragraph{Self- and Weakly-Supervised Learning}
	
	Self and weakly-supervised approaches tackle the generalization problem by learning a meaningful representation from unlabeled samples in other domains. The supervision stems from unpaired 2D and 3D body pose annotations \cite{wandt2019repnet,drover20183d} or from multi-view images \cite{bouazizi2021self,Kocabas19}. Kocabas \textit{et.~al.}~\cite {Kocabas19} triangulate 2D pose estimates in a multi-view environment to generate pseudo-labels for 3D body pose training. Wandt \textit{et.~al.}~\cite{wandt2019repnet} propose the re-projection network to learn the mapping from 2D to the 3D body pose distribution using adversarial learning. In particular, the critic network improves the generated 3D body pose estimate based on the Wasserstein loss~\cite{arjovsky2017wasserstein} and unpaired 2D and 3D body poses. Similarly, Drover~\textit{et.~al.}~\cite{drover20183d} rely on a discriminator network for supervision of 2D body pose projections. However, the method additionally utilizes 3D ground-truth data to generate synthetic 2D body joints during the training. Kundu ~\textit{et.~al.}~\cite{kundu2020self} present a self-supervised 3D pose estimation approach with an interpretable latent space. To better generalize across scenes and datasets, the approach still relies on unpaired 3D poses. Tripathi \textit{et.~al.} ~\cite{tripathi2020posenet3d} propose a method to regress the 3D keypoints by incorporating the temporal information next to the adversarial objective. A network distillation is employed for additional supervision.  Instead of adversarial learning with unpaired 3D poses, we rely on a multi-view system to reach the same goal. Different from all these methods, we are the first to leverage multi-view video sequences via self-supervised learning. In Tab.~\ref {tab:comparison} we give  a characteristic comparison of our approach against prior work.
	

	\section{Method}
	
	We present our approach to infer the 3D body pose of a single person in video sequences. Instead of performing 3D body pose estimation directly on each image frame, we extract 2D body pose estimates $\mathbf{y}_{0}, \dots, \mathbf{y}_{T}$ over time $T$, which composes the input to our approach. Our goal is to regress the 3D body pose $\mathbf{Y}_{0}, \dots, \mathbf{Y}_{T}$ for each 2D pose estimate, where the 3D body pose $\mathbf{Y}_{t} \in \mathbb{R}^{3 \times N}$ at the time step $t \in T$ consists of $N$ joints. We describe the 2D-to-3D body pose mapping as: 
	\begin{equation}
		\mathbf{Y}_{0}, \dots, \mathbf{Y}_{T} = f(\mathbf{y}_{0}, \dots, \mathbf{y}_{T};\mathbf{\theta})
	\end{equation}
	where $f:\mathbb{R}^{2 \times N \times T} \rightarrow \mathbb{R}^{3 \times N \times T}$ corresponds to the mapping function. We approximate the mapping function with a convolutional neural network that is parametrized by $\mathbf{\theta}$. Learning the model parameters normally is performed with ground-truth information. In this work, we propose to learn the model parameters with supervision, which we derive from a multiple-view camera system and 2D body pose estimates.

\begin{figure}[t]
		\centering
		
		\includegraphics[width=\linewidth]{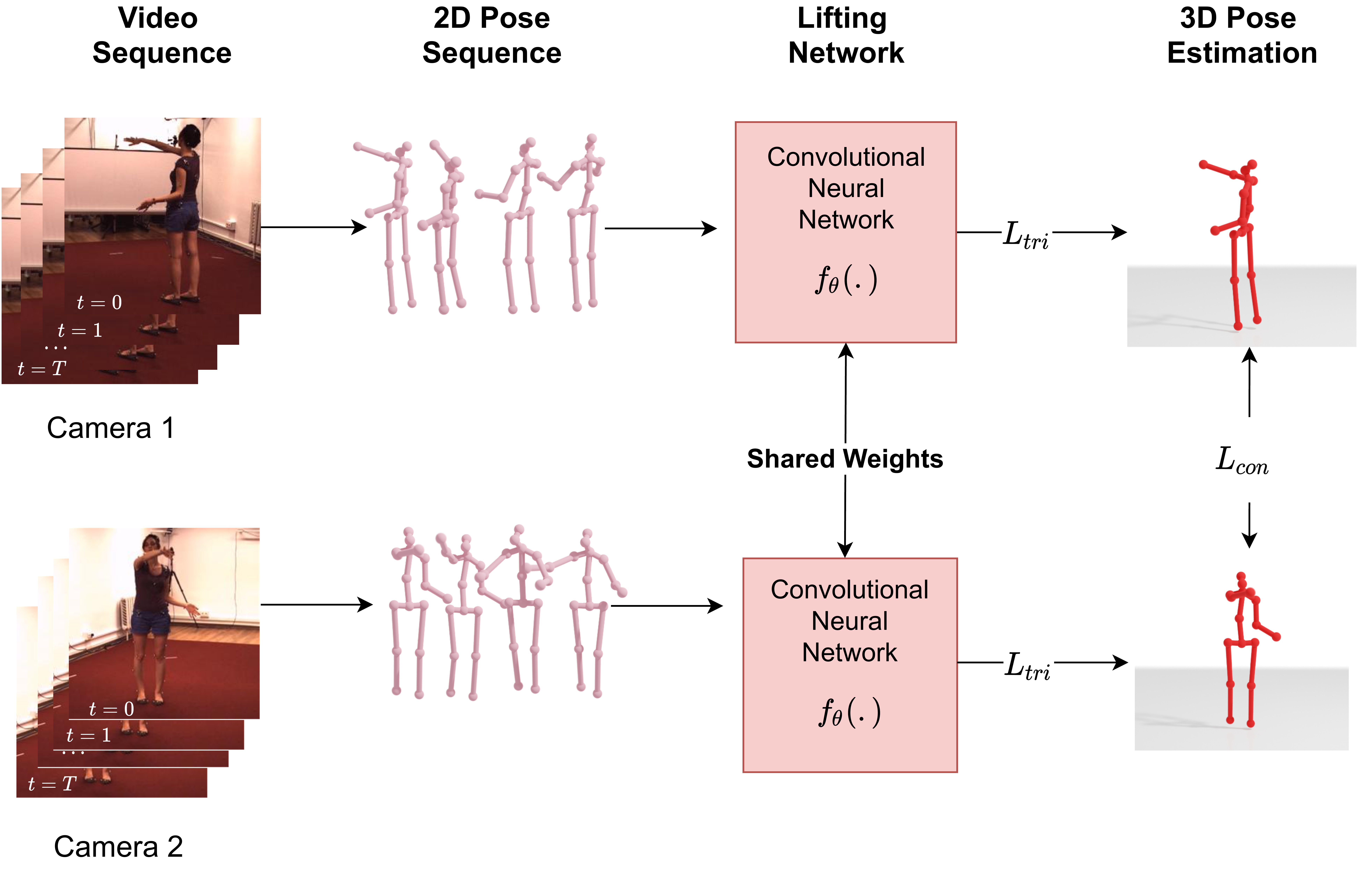}
		\caption{Illustration of our learning algorithm.  During training, we optimize the lifting network with dilated temporal convolution $f_{\mathbf{\theta}} (\cdot)$ to map a sequence of 2D pose estimate to 3D body poses with geometric loss functions. As input to our model we assume multiple view 2D pose detections as shown on the left. First we compute the triangulated 3D pose from the 2D detections as pseudo-labels for the input triangulation loss $\mathcal{L}_{tri}$. The Multi-view consistency loss $\mathcal{L}_{con}$ is then used to enforce that the estimated keypoints from both views can be transformed to each other via the known rigid transformation. Both losses  are then combined to train the 3D pose network $f_{\mathbf{\theta}}$. During inference, our approach is only single-view. Note that our approach targets single person 3D body pose estimation. }
		\label{fig:method}
	\end{figure}

	We assume having access to a multiple-view time-synchronized system with $C$ cameras and an off-the-shelf 2D body pose detector. During training, a 2D pose sequence $ (\mathbf{{y}}_{c,0}, \dots, \mathbf{{y}}_{c,T})$ from a camera $c$ is subsequently fed to the convolutional neural network   $f_{\mathbf{\theta}}$ to predict the 3D body poses. To constrain the three-dimensional search space of possible poses, we make use of multiple-view geometry to obtain pseudo 3D body poses for $\mathcal{L}_{tri}$. The triangulation loss $\mathcal{L}_{tri}$ minimizes the difference between the 3D body pose predictions $\mathbf{{Y}}$ and triangulated poses $\mathbf{\hat{Y}}_{tri}$ (set as ground-truth). To ensure a view consistency, we rely on the geometric consistency loss $\mathcal{L}_{con}$, which enforces the network to learn poses that are view invariant. The estimated keypoints from two different views can be transformed to each other via a known rigid transformation (\textit{Translation and Rotation}). The proposed learning approach is then self-supervised with the geometric loss functions, namely the input triangulation $\mathcal{L}_{tri}$ and the geometric consistency $\mathcal{L}_{con}$. Fig.~\ref{fig:method} shows the overall framework of our proposed method. Note that our learning algorithm makes use of all available camera views during training, while the inference is single-view. Next, we present the motivation and elements of the proposed loss functions in detail.

	\subsection{Triangulation Loss}

	We make use of the Direct Linear Triangulation (DLT)~\cite {Cambridge2004} method for obtaining the 3D keypoint position from the 2D body pose estimates. We apply the same approach to all body pose landmarks. Similar to \cite{Kocabas19}, we consider the obtained 3D human pose estimate $\mathbf{\hat{Y}}_{tri}$ as ground-truth. The input triangulation loss is defined by:
	\begin{equation}\label{input_loss}
		\mathcal{L}_{tri} = \sum_{c=1}^{C} \sum_{t=1}^{T}  \parallel \tau_{w \rightarrow c}(\mathbf{\hat{Y}}_{tri,t})  - f_{\mathbf{\theta}}(\mathbf{{y}}_{c,t}) \parallel_2,
	\end{equation}
	where $\tau_{w \rightarrow c} (\cdot)$ corresponds to the transformation from the world coordinate system  $w$ to the camera coordinate system $c$. It is given by:
	\begin{equation}\label{world_camera}
		\mathcal{\tau}_{w \rightarrow c} = \mathcal {R}_{w \rightarrow c} (\mathbf{\hat{Y}}_{tri}- \mathcal {T}_{w \rightarrow c}),
	\end{equation}
	where  $\mathcal {R}_{w \rightarrow c}  \in \mathbb{R}^{3\times3}$ denote the rotation matrix and $\mathcal {T}_{w \rightarrow c}  \in \mathbb{R}^{3\times1}$ for the corresponding translation vector. Since the triangulation loss highly depends on the quality of the detected landmarks for each camera view, relying only on the input triangulation would make the 3D body poses fixed during the training. The errors of 3D reconstruction will then directly propagate to the network $f_{\mathbf{\theta}}$. To this end, we propose to transform each predicted pose to another camera view using the geometric consistency loss, as presented below.

	\subsection{Geometric Consistency Loss}

	The goal of the geometric consistency loss is to ensure that the predicted 3D body pose is consistent across different views. Specifically, the 3D body pose $\mathbf{{Y}}$, when accordingly rotated and translated should be consistent with the corresponding pose in the second view, regardless of the 2D body pose input. Based on this observation, we use the consistency loss that is given by: 
	\begin{equation}\label{cons_loss}
		\mathcal{L}_{con} =  \sum_{c=1}^{C} \sum_{\substack{c\prime=1 \\ c \neq c\prime}}^{C} \sum_{t=1}^{T}  {\parallel f_{\mathbf{\theta}}(\mathbf{{y}}_{c,t}) - \tau_{c\prime \rightarrow c}(f_{\mathbf{\theta}}(\mathbf{{y}}_{c\prime,t})) \parallel }_2
	\end{equation}
	where $\tau_{c\prime \rightarrow c} (\cdot)$ corresponds to the transformation from the camera $c\prime$ to the camera $c$ coordinate system. It is defined as:
	\begin{equation}\label{cons_transformation}
		\mathcal{\tau}_{c\prime \rightarrow c} = \mathcal {R}_{c\prime \rightarrow c} f_{\mathbf{\theta}}(\mathbf{{y}}_{c,t})- \mathcal {T}_{c\prime \rightarrow c}.
	\end{equation}
	Moreover, the camera transformation is given by:
	\begin{equation}\label{rotation}
		\mathcal {R}_{c\prime \rightarrow c} = \mathcal {R}_{c}  \mathcal {R}_{c\prime}^\top \text{ and }  \mathcal {T}_{c\prime \rightarrow c} = \mathcal {R}_{c}   (\mathcal{T}_{c\prime} - \mathcal{T}_{c}).
	\end{equation}
	
	\subsection{Complete Objective}
	
	With the geometric consistency loss, the model learns to predict body poses that are robust to camera view changes. Due to the fact that the ground-truth 3D body pose is unknown during training, the multi-view consistency is used as a form of self-supervision. Enforcing only multi-view consistency is not sufficient to infer accurate 3D body poses across different camera views, since it will lead to a degenerate solution where all keypoints collapse to the same pose. To this end, both the triangulation loss $\mathcal{L}_{tri}$ and $\mathcal{L}_{con}$ are used to train the network $f_{\mathbf{\theta}}$. To learn the parameters $\mathbf{\theta}$, we train our model based on the proposed loss functions and the training samples from all camera views. We obtain the model parameters by minimizing the following objective:

	\begin{equation}
		\label{training_eq}
		\theta^{\prime} = \arg \min_{\theta} \mathcal{L}_{tri}  + \mathcal{L}_{con} ,
	\end{equation} 
	
	A summary of the training of our method is illustrated in Algorithm~\ref{alg:ttt}.
	\begin{algorithm}[h]\small
		\caption{3D Human Pose Estimation Training Algorithm using Temporal Information and Multi-view Geometry. }
		\label{alg:ttt}
		\SetKwInOut{Input}{Input} \SetKwInOut{Output}{Output}
		\Input{2D pose estimate  $(\mathbf{y}_{0}, \dots, \mathbf{y}_{T})$, learning rate $\alpha$, number of samples $\mathbf{S}_{c}$, cameras $c$ and $c\prime$.} 
		\Output{3D pose estimations $(\mathbf{Y}_{0}, \dots, \mathbf{Y}_{T})$.}
		
		\For{ $epoch$ $<$ $epoch_{max}$ }{
			\For{ $i = 1$ to $\mathbf{S}_{c}$ }{
				
				Compute
				$\nabla_{\boldsymbol{\theta}}\mathcal{L}_{tri}(\mathbf{\hat{Y}}_{c,i},\boldsymbol{Y}_{c,i};\boldsymbol{\theta}_i)$ \\
				
				Compute
				$\nabla_{\boldsymbol{\theta}}\mathcal{L}_{con}(\mathbf{\hat{Y}}_{c \rightarrow c\prime,i},\boldsymbol{Y}_{c\prime,i};\boldsymbol{\theta}_i)$ \\
				
				Compute complete objective
				
				$\nabla_{\boldsymbol{\theta^{\prime} }}\mathcal{L}_{s}(\mathbf{\hat{Y}}_{i},\boldsymbol{Y}_i;\boldsymbol{\theta}^{\prime} _i)$ using Equation \ref{training_eq}

				Update  parameters: $\boldsymbol{\theta}^{\prime}_i \leftarrow \boldsymbol{\theta}^{\prime}_i-\alpha \nabla_{\boldsymbol{\theta}}\mathcal{L}_{s}(\mathbf{\hat{Y}}_{i},\boldsymbol{Y}_i;\boldsymbol{\theta}^{\prime}_i)$ .
			}
		}
	\end{algorithm}

	
	\section{Experiments}
	
	\subsection{Experimental Setup}
	
	We quantitatively evaluate our method on two 3D human pose estimation benchmarks, namely the Human3.6M ~\cite{ionescu2013human3} and MPII-INF-3DHP ~\cite{mpiiinfdataset}. For Human3.6M, we train our model on five subjects (S1, S5, S6, S7 and S8) and evaluate on two subjects (S9 and S11). We use three evaluation protocols: Protocol 1 refers to the Mean Per Joint Position Error (MPJPE). Protocol 2 results in the Mean Per Joint Position after procrustes alignment to the ground-truth 3D poses by a rigid transformation (PMPJPE) and Protocol 3 aligns the predicted poses with the ground-truth only in scale (N-MPJPE). For MPI-INF-3DHP ~\cite{mpiiinfdataset}, which is a recently published dataset with $8$ actors performing $8$ actions, we also follow the standard protocol ~\cite{mpiiinfdataset}: The five chest-height cameras, which provide 17 joints (compatible with Human3.6M ~\cite{ionescu2013human3}) are used for training.  For evaluation, we use the official test set which includes challenging outdoor scenes. We report the results by means of 3D Percentage of Correct Keypoints (PCK) with a threshold of $150mm$ and the corresponding Area Under Curve AUC so that we are consistent with \cite{wandt2019repnet,mpiiinfdataset,habibie2019wild,kolotouros2019learning}. 
	
	\subsection{Implementation Details}
	Our  network is a fully convolutional architecture with residual connections and dilated convolutions ~\cite{Pavllo19}. Unlike recurrent structures that do not support parallelization over time and tend to drift over long sequences ~\cite{Pavllo19}, dilated temporal convolutions are computationally efficient and maintain the long-term coherence. We choose four different frame sequence lengths when conducting our experiments, i.e.$f = 1$, $f = 27$, $f = 81$, $f = 243$. The influence of the number of frames is discussed in section \ref{Ablation studies}. Pose flipping is applied as data augmentation during training. We train our model using the Adam optimizer for $60$ epochs with weight decay of $0.1$. An exponential learning rate decay schedule with the initial learning rate of $2e^{-4}$ and decay factor of $0.98$ after each epoch is adopted. The batch size is set to $1024$. As a 2D pose detector, we used, following \cite{Pavllo19} the cascaded pyramid network (CPN) \cite{chen2018cascaded}.

	
	\begin{table*} []  
		\begin{center}
			\caption{Results on the Human3.6M dataset. Comparison of our self-supervised approach with state-of-the-art weakly- and fully-supervised methods following evaluation Protocol-I (without rigid alignment) individually for all 15 actions. All values are given in $mm$.}
			\label{tab:h36m_mpjpe}
			\resizebox{\textwidth}{!}{\begin{tabular}{l l| l | c c c c c c c c c c c c c c c | c}
					&Supervision& Method & Dir. & Dis. & Eat & Greet & Phone & Photo & Pose
					& Purch. & Sit & SitD & Smoke & Wait & WalkD & Walk
					& WalkT & Avg\\
					\hline
					&& Zhou et al. \cite {zhou2016deep} & 91.8 &  102.4 & 96.7 & 98.8 & 113.4  & 125.2 & 90.0 & 93.8 & 132.2  & 159.0 &  107.0 & 94.4 & 126.0 & 79.0 & 99.0 & 107.3 \\
					&& Lu et al. \cite {Lu2018} & 68.4 &  77.3 & 70.2 & 71.4 & 75.1  & 86.5 & 69.0 & 76.7 & 88.2 & 103.4 &  73.8 & 72.1 & 83.9 & 58.1 & 65.4 & 76.0 \\
					&Full& Pavlakos et al. \cite{pavlakos2017coarse} & 67.4 &  71.9  & 66.7 & 69.1 & 72.0  & 77.0 & 65.0 & 68.3  & 83.7 & 96.5 &  71.7  & 65.8 & 74.9 & 59.1 & 63.2 & 71.9 \\
					&& Martinez et al. \cite {Martinez17} & 51.8 &  56.2 & 58.1  & 59.0 & 69.5  & 78.4 & 55.2 & 58.1 & 74.0 & 94.6 & 62.3  & 59.1 & 65.1  & 49.5 & 52.4  & 62.9 \\
					&& Pavllo et al. \cite {Pavllo19} & 45.1  &  47.4 & 42.0 & 46.0 & 49.1  & 56.7 & 44.5 & 44.4  & 57.2 & 66.1 & 47.5 & 44.8 & 49.2 & 32.6 & 34.0 & 47.1 \\[0.3em]
					\hline 
					\hline
					
					&& Li et al.  \cite {li2020weakly}   & 62.0 &  69.7 & 64.3   & 73.6 & 75.1   & 84.8 & 68.7  & 75.0 & 81.2   & 104.3 & 70.2  & 72.0 &  75.0 & 67.0 & 69.0  & 73.9 \\
					&Weak& Wandt et al. \cite {wandt2019repnet}& 77.5  &  85.2 & 82.7 & 93.8 & 93.9  & 101.0 & 82.9 & 102.6  & 100.5 & 125.8 & 88.0 & 84.8 & 72.6 & 78.8 & 79.0 & 89.9 \\[0.3em]
					\hline 
					&& Ours (self-supervised)  & 48.2 & 49.3 & 46.5 & 48.4 & 52.4 & 46.5 & 46.4
					& 61.4 & 72.3 & 51.0 & 59.8 &  46.7 & 37.5 & 52.1 & 39.1 & \textbf{50.6} \\

					\bottomrule
			\end{tabular}}
		\end{center}
	\end{table*}
	
	\begin{table*} [] 
		\begin{center}
			\caption{Results on the Human3.6M dataset. Comparison of our self-supervised approach with state-of-the-art supervised and weakly- and self-supervised methods following evaluation Protocol-II (with rigid alignment) individually for all 15 actions. All values are given in $mm$. }
			\label{tab:h36m_pmpjpe}
			\resizebox{\textwidth}{!}{\begin{tabular}{l l | l | c c c c c c c c c c c c c c c | c}
					&Supervision&Method & Dir. & Dis. & Eat & Greet & Phone & Photo & Pose
					& Purch. & Sit & SitD & Smoke & Wait & WalkD & Walk
					& WalkT & Avg\\
					\hline
					
					&& Zhou \etal~ \cite {zhou2016sparse} & 99.7 & 95.8 & 87.9 & 116.8 & 108.3  & 107.3 & 93.5 & 95.3 & 109.1 & 137.5 &  106.0 & 102.2 & 110.4 & 106.5 & 115.2 & 106.7 \\
					&& Bogo \etal~ \cite {bogo2016smpl} & 62.0 &  60.2 & 67.8 & 76.5 & 92.1  & 77.0 & 73.0 & 75.3 & 100.3 & 137.3 &  83.4 & 77.3 & 79.7 & 48.0 & 87.7 & 82.3 \\
					&Full& Martinez \etal~ \cite {Martinez17} & 39.5 &  43.2 & 46.4 & 47.0 & 51.0  & 56.0 & 41.4 & 40.6 & 56.5 & 69.4 & 49.2 & 45.0 & 38.0 & 49.0 & 43.1 & 47.7 \\
					&& Lu \etal~ \cite {Lu2018} & 40.8 &  44.6 & 42.1 & 45.1 & 48.3  & 54.6 & 41.2 & 42.9 & 55.5 & 69.9 &  46.7 & 42.5 & 36.0 & 48.0 & 41.4 & 46.6 \\
					&& Pavllo \etal~ \cite {Pavllo19} & 34.2  &  36.8 & 33.9 & 37.5 & 37.1  & 43.2 & 34.4 & 33.5 & 45.3 & 52.7 & 37.7 & 34.1 & 38.0 & 25.8 & 27.7 & 36.8 \\[0.3em]
					\hline
					\hline
					
					&& Wu \etal~ \cite {Wu_2016} & 78.6 &  90.8 & 92.5 & 89.4 & 108.9 & 112.4 & 77.1 & 106.7 & 127.4 & 139.0 & 103.4 & 91.4 & 79.1 & - & - & 98.4 \\
					&& Tung \etal~ \cite{tung2017adversarial} & 77.6 &  91.4 & 89.9 & 88.0 & 107.3 & 110.1 & 75.9 & 107.5 & 124.2 & 137.8 &  102.2 & 90.3 & 78.6  & - & - & 97.2\\  
					&Weak& Wandt \etal~ \cite {wandt2019repnet} & 53.0 &  58.3 & 59.6 & 66.5 & 72.8 & 71.0 & 56.7 & 69.6 & 78.3 & 95.2 & 66.6 & 58.5 & 63.2 & 57.5 & 49.9 & 65.1 \\
					&&  Zhou \etal~ \cite {zhou2017towards} & 54.8 &  60.7 & 58.2 & 71.4 & 62.0 & 65.5 & 53.8 & 55.6 & 75.2 & 111.6 & 64.1 & 66.0 & 50.4 & 63.2 & 55.3 & 64.9 \\
					
					&& Drover \etal~ \cite {drover20183d} & 60.2 &  60.7 & 59.2 & 65.1 & 65.5 & 63.8 & 59.4 & 59.4
					& 69.1 & 88.0 &  64.8 & 60.8 & 64.9  & 63.9 & 65.2 & 64.6\\[0.3em]
					\hline
					&& Chen \etal~ \cite {chenunsupervised19} & 55.0 &  58.3 & 67.5 & 61.8 & 76.3 & 64.6 & 54.8 & 58.3 & 89.4 & 90.5 & 71.7 & 63.8 & 65.2 & 63.1 & 65.6 & 68.0 \\ 
					
					&Self& Kocabas \etal~ \cite{Kocabas19} & - &  - & - & - & - & - & - & - & - & - & - & - & - & - & - & 67.5 \\ 
					&&  Bouazizi \etal~\cite{bouazizi2021self} & 49.4 &   51.7 &  61.7 &  56.5 & 64.9 &  67.1 &  51.6 & 52.1 &  83.9 &   111.3 & 60.5  & 54.7 &  56.9  & 45.9  & 53.6 & 62.0 \\ 
					&& Tripathi \etal~ \cite {tripathi2020posenet3d} & 49.1 &  52.4 & 57.5 & 56.4 & 63.5 & 59.5 &  51.3 & 48.4 & 77.1 &  81.5 & 60.4  & 59.6 & 53.5  & 59.1  & 51.4 & 59.4 \\

					\hline
					
					&& Ours (self-supervised)  & 37.1 & 38.4 & 38.2 & 39.7 & 40.9 & 36.3 & 35.2
					& 49.4 & 59.2 & 40.9 & 46.3 &  36.5 & 29.6 & 40.6 & 31.3 & \textbf{40.0} \\
					
					\bottomrule
			\end{tabular}}
		\end{center}
	\end{table*}
	
	\subsection{Human3.6M Evaluation}

\begin{figure}[]
		\centering    
		\subfigure[\textit{Directions S9 frame 1, 100, 200, 300, 400, 500, 600, 700.} ]{\label{fig:h}\includegraphics[width=\linewidth]{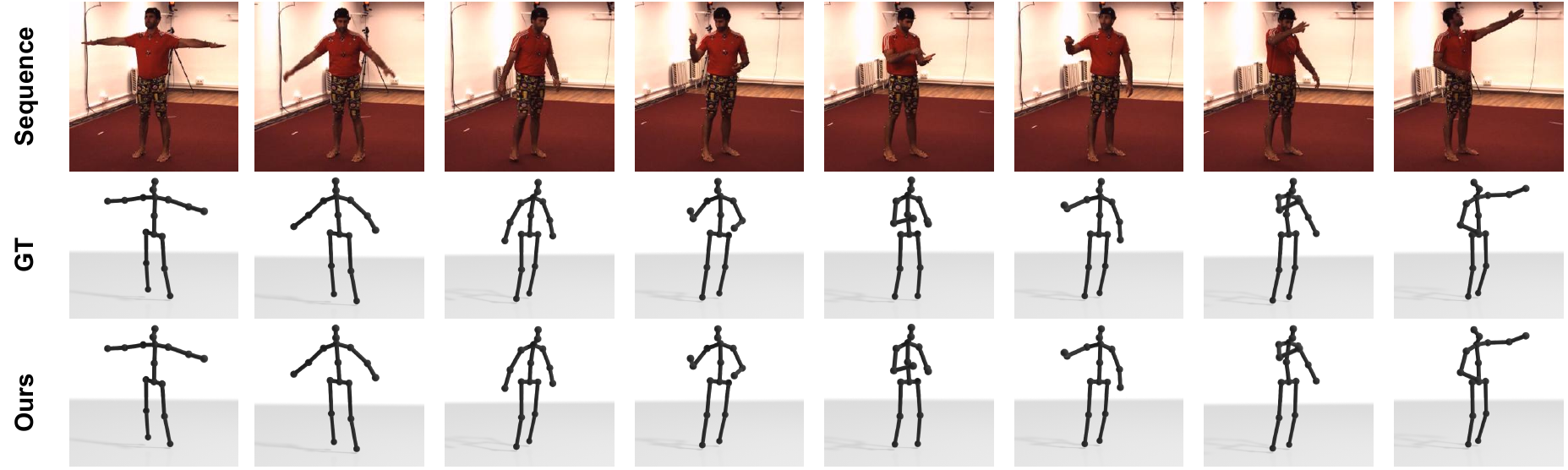}} 
		
		\subfigure[\textit{WalkDog S9 frame 200, 300, 400, 500, 600, 700, 800, 900.} ]{\label{fig:b}\includegraphics[width=\linewidth]{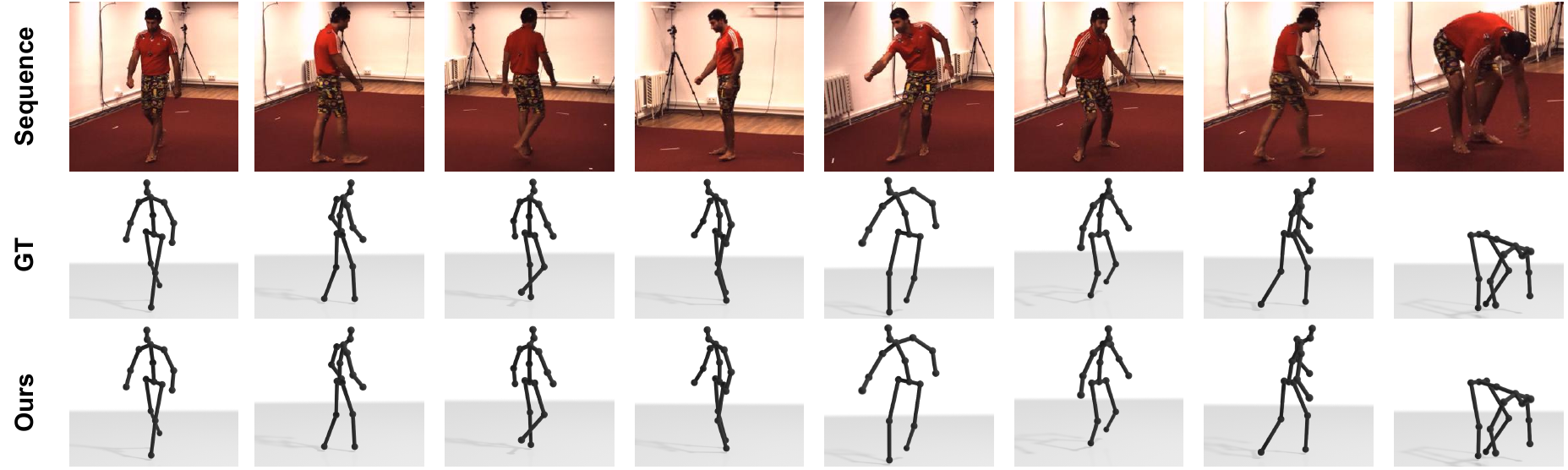}}
		
		\subfigure[\textit{Greeting S11 frame 1, 200, 400, 600, 800, 1000, 1200, 1400.}]{\label{fig:b}\includegraphics[width=\linewidth]{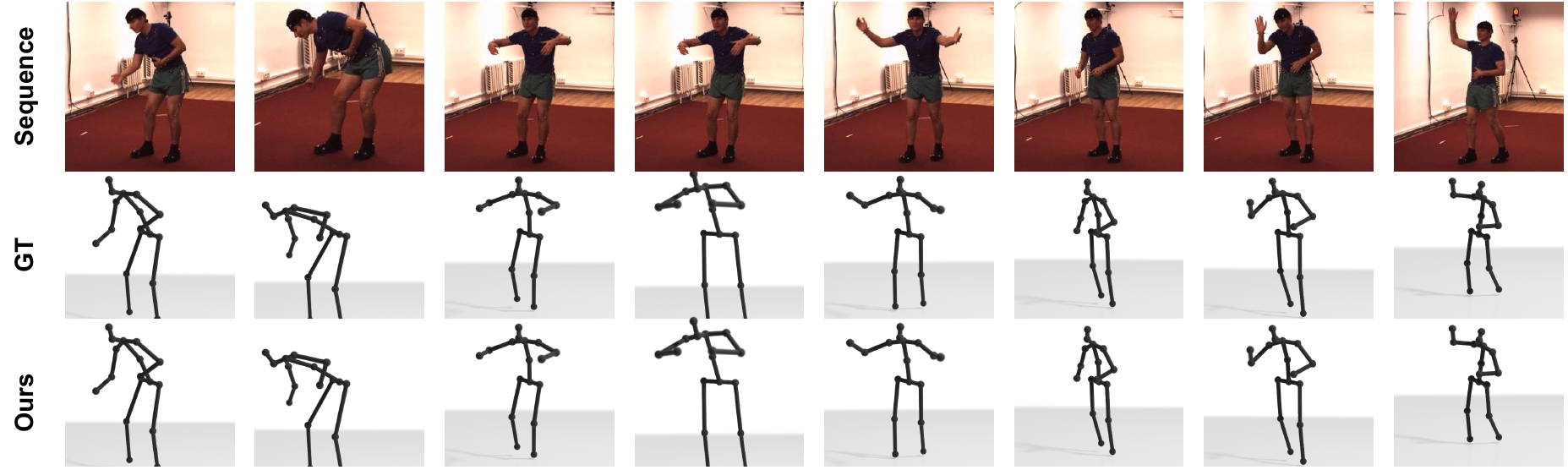}}

		\caption{ Qualitative results on Human3.6M dataset \\ 
			\textbf{Top:} Video Sequence \textbf{Middle:} 3D Ground-truth \textbf{Bottom:} 3D Pose Reconstruction }

		\label{h36m_qualitative}
	\end{figure}

	We first report the result of all single $15$ action on the Human3.6M dataset and compare with state-of-the-art  approaches. The results in Tab.~\ref{tab:h36m_mpjpe} and \ref{tab:h36m_pmpjpe}, show that our self-supervised method outperforms all weakly and self-supervised methods by a large margin. Our approach also compares favorably to the state-of-the-art fully-supervised approaches. It achieves an MPJPE of $50.6 mm$, which is only $3 mm$ higher than the supervised approach from \cite{Pavllo19} using 3D ground-truth keypoints and $26 mm$ better than \cite{Kocabas19} that is relying only  on multi-view geometry. For protocol 2, we also obtain the best overall result of $40.0 mm$ as shown in Tab.~\ref{tab:h36m_pmpjpe}. This clearly demonstrates the advantage of incorporating the temporal information over single-frame approaches. Compared to the self-supervised approach of \cite{tripathi2020posenet3d}, which makes use of unpaired 3D pose keypoints next to the temporal information, our approach still yields an error reduction of $33.2 \%$. In addition, our method reaches more accurate pose predictions than the fully-supervised approach of Pavallo \textit{et.~al.} ~\cite{Pavllo19} on difficult actions like \textit{SittingDown, WalkDog, Photo}. Fig.~\ref{h36m_qualitative} provides a visual comparison between the predicted poses and the ground-truth 3D body poses. We evaluate our self-supervised approach on the three challenging actions \textit{Directions, WalkDog, Greeting} and show that our network is able to infer the 3D body poses in an accurate way.

	\subsection{MPII-INF-3DHP Evaluation}

\begin{figure} []
		\centering    
		\subfigure[\textit{Test sequence1 frame 1, 200, 400, 600, 800, 1000, 1200, 1400.}]{\label{fig:a}\includegraphics[width=\linewidth]{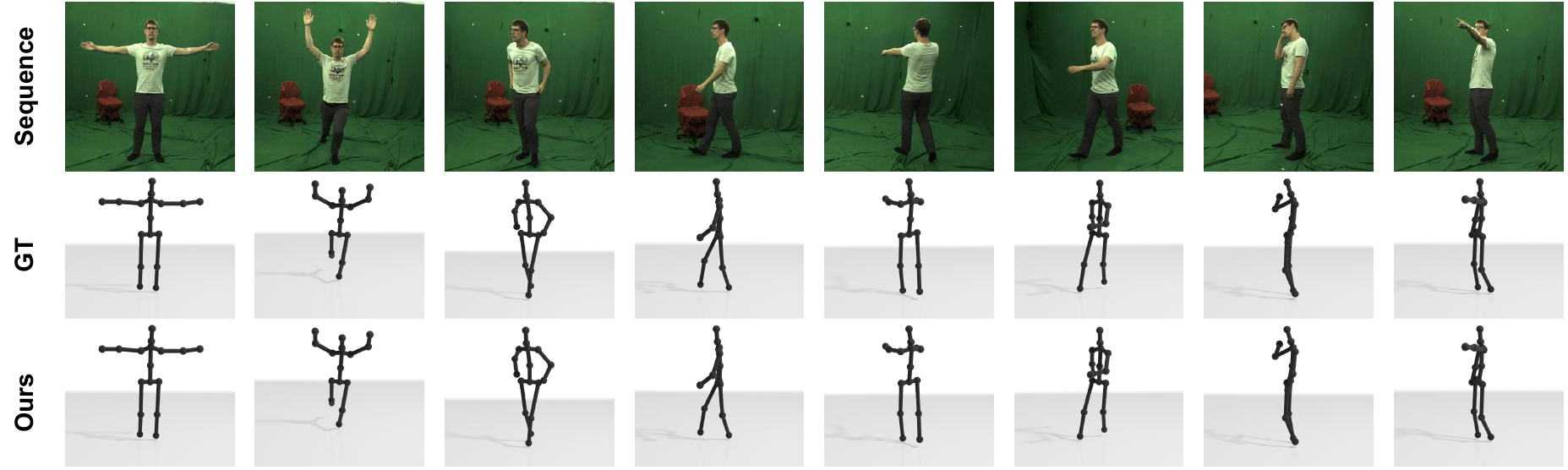}}
		
		\subfigure[\textit{Test sequence2 frame 1, 200, 400, 600, 800, 1000, 1200, 1400.}]{\label{fig:b}\includegraphics[width=\linewidth]{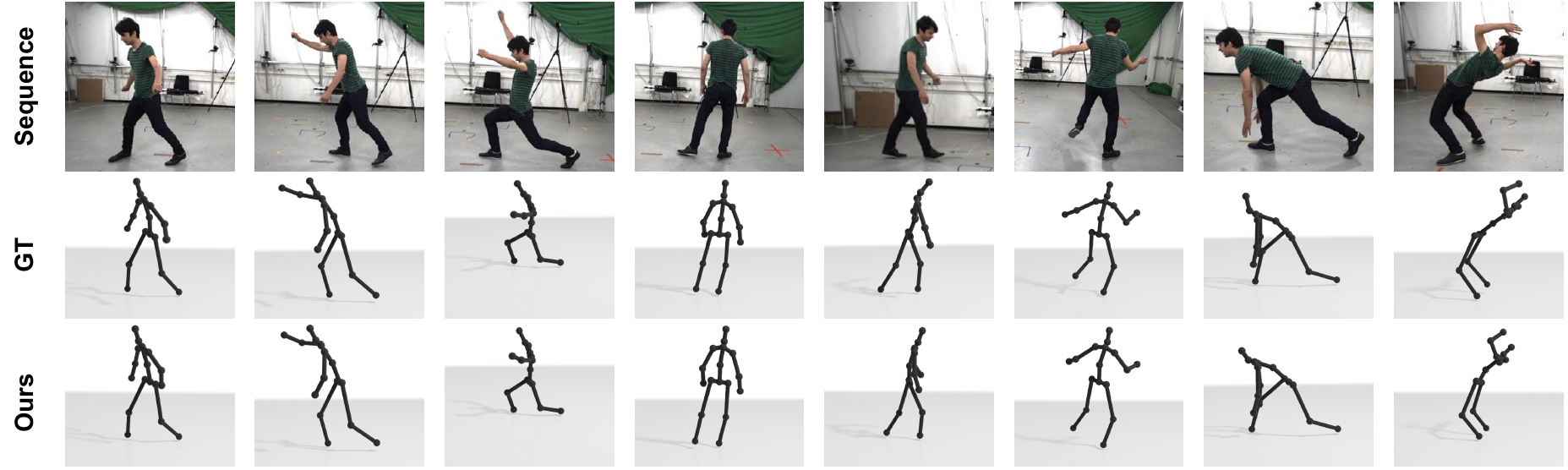}}
		
		\subfigure[\textit{Test sequence3 frame 1, 200, 400, 600, 800, 1000, 1200, 1400.}]{\label{fig:b}\includegraphics[width=\linewidth]{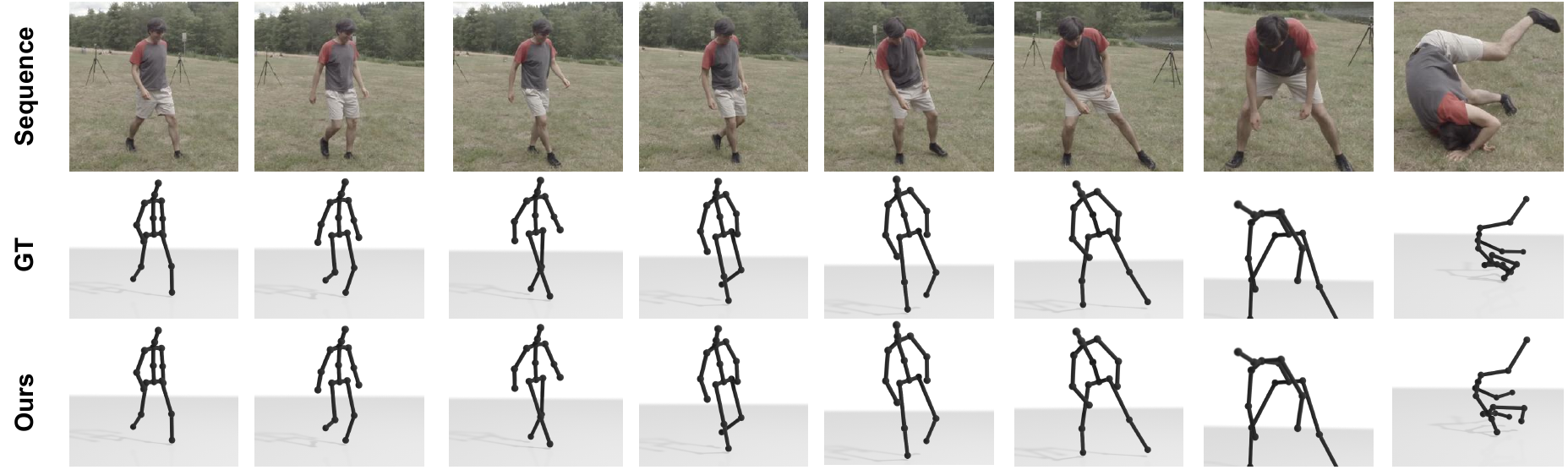}}
		
		\caption{ Qualitative results on MPII-INF-3DHP dataset \textbf{Top:} Video Sequence \textbf{Middle:} 3D Ground-truth \textbf{Bottom:} 3D Pose Reconstruction }
		\label{mpii_qualitative}
	\end{figure}

	We report the results on the MPII-INF-3DHP dataset in Tab.~\ref{tab:eval_3dhp} and compare to other state-of-the-art approaches. For a fair comparison, our model utilizes the same 2D pose keypoints as in \cite{Kocabas19}. Our method significantly outperforms all weakly and self-supervised approaches across different evaluation metrics and obtains $30\%$ better PCK than \cite{Kocabas19} that also uses multiple-view geometry. This clearly demonstrates the advantage of incorporating the temporal information in a multi-view setting. Our method yields the lowest  average  PMPJPE  of $51.1 mm$ as  shown in Tab.~\ref{tab:eval_3dhp}, which is comparable to most state-of-the-art method. We also quantitatively evaluate the performance on MPII-INF-3DHP given a model trained on Human3.6M. Without any training or fine-tuning, our model reaches a PCK of $74\%$, which is better than previous weakly- and self-supervised approaches \cite{rhodin2018learning, kanazawa2018end, habibie2019wild,kolotouros2019learning, iqbal2020weakly, Kocabas19,chenunsupervised19, li2020geometry} trained on this dataset. These results suggest that the generalization is significantly improved by incorporating the temporal information and the self-consistency reasoning. To further demonstrate the effectiveness of our training strategy, we qualitatively evaluate our approach on constrained indoor scenes and complex outdoor scenes, covering a notable diversity of poses. As shown in the Fig.~\ref{mpii_qualitative}, our approach is able to generalize across unseen poses, appearances and subjects.

	\subsection{Ablation Study}
	\label{Ablation studies}
	To verify the contribution of the individual components of our method on the performance, we conduct extensive ablation experiments on the Human3.6M dataset.

	\textbf{Impact of the 2D Pose Estimation} Since the performance of the lifting network highly depends on the 2D pose estimator, we evaluate different 2D detectors that do not rely on ground-truth bounding boxes and report the results in Tab.~\ref{tab:detector}. The performance with \textit{Cascaded Pyramid Network} CPN  \cite{chen2018cascaded} is about $5 mm$ better than the model trained with \textit{Detectron} \footnote{\url{https://github.com/facebookresearch/Detectron/}} landmarks. To further investigate the lower bound of our method, we directly use the ground-truth 2D keypoints as input to alleviate error caused by noisy detections. Following \cite{Martinez17,wandt2019repnet}, we then add different levels of gaussian noise $\mathcal{N}(0,\sigma)$ to the ground-truth 2D keypoints and report the results under protocol 2  in Tab.~\ref{tab:noise}. The results indicate that the noise has a major impact on the model performance.

	\textbf{Impact of the  multi-view information} We show the results for the training with different number of cameras in Tab.~\ref {tab:ablations}. Since multiple views are observed and the temporal information is not discarded leading to more accurate body posture and constant bone lengths, our method yields the lowest average PMPJPE of $50.60  mm$ with four cameras. While the performance slightly drops due to the lower number of training samples and views, our approach still produces promising results which enables the use of our model in the-wild.  We further illustrate the ablation study on the consistency constraint. We see a $2.1\%$ error drop ($51.4 mm \rightarrow 50.60 mm$), when removing the consistency constraint. This shows that the  geometric consistency reasoning indeed improve the 3D body pose predictions.
	
	\textbf{Impact of the receptive field } To verify the impact of the receptive field, we choose four different frame sequence lengths when conducting our experiments. Since the temporal dimension of video sequences encodes valuable information, the lowest average error is achieved with the largest receptive field. Nevertheless we still achieve very promising results with $f = 27$ and $f = 81$. This clearly demonstrates the advantage of incorporating the temporal information, which encourages smooth motions and alleviates  drift over long sequences .

	
	\begin{table} [htbp]
		\footnotesize
		\caption{Evaluation results for the MPII-INF-3DHP dataset. Best results are marked in bold. MPJPE and PMPJPE are given in $mm$, PCK and AUC are given in \%. Best in bold, second best underlined.}
		\centering
		\setlength\tabcolsep{3pt}
		\begin{tabular}{ l | l | c c c c}
			Supervision & Method & MPJPE$\downarrow$ & PMPJPE$\downarrow$ & PCK$\uparrow$ & AUC$\uparrow$ \\
			\hline
			
			Full& Habibie \cite{habibie2019wild} & 127.0 & 92.0 & 69.6  & 35.5 \\
			& HMR \cite{kanazawa2018end} & 124.2 & 86.3 & 72.9 & 36.5  \\
			& Kolotouros \cite{kolotouros2019learning} & 105.2 & 67.5 & 76.4  & 37.1 \\
			\hline
			\hline
			
			weak            & Rhodin \cite{rhodin2018learning}          & 121.8 & - & 72.7  & -\\
			& HMR \cite{kanazawa2018end}             & 169.5 & 113.2 & 59.6 & 27.9 \\
			& Habibie \cite{habibie2019wild} & 127.0 & 92.0 & 66.8  & 35.5 \\
			& Kolotouros \cite{kolotouros2019learning}          & 124.8 & 80.4& 66.8 & 30.2 \\
			& Iqbal \cite{iqbal2020weakly}             & 110.1 & - & 76.5 & -  \\
			
			\hline
			self
			
			& Kocabas  \cite{Kocabas19}  & 125.7 & - & 64.7  & - \\ 
			& Bouazizi \cite{bouazizi2021self}               & -  &  - & 65.9  & 32.5 \\
			& Chen  \cite{chenunsupervised19}               & -  &  - & 71.1  & 36.3 \\
			& Li \cite{li2020geometry}                    & -     & - & 74.1  & 41.4\\
			& Kundu \cite{kundu2020self}              & 103.8 & - & \textbf{82.1}  & \textbf{56.3}\\
			\hline
			& Ours   \textit{(H36M)} & 114.7 & 75.4 & 74.1 & 38.8 \\
			
			& Ours   \textit{(3DHP)} & \textbf{93.0}  & \textbf{51.1} &  \underline{81.0} &  \underline{50.1} \\

			\bottomrule
		\end{tabular}
		\label{tab:eval_3dhp}
	\end{table}

	
	\begin{table} [h!tp]
		\footnotesize
		\caption{Effect of the multi-view information. }
		\centering
		\begin{tabular}{ l | c c c }
			& MPJPE$\downarrow$     & PMPJPE$\downarrow$ & NMPJPE$\downarrow$ \\
			\hline
			2 cameras                  & 60.00 & 45.20 & 57.40 \\
			3 cameras                & 52.70 & 41.10 & 51.10 \\
			4 cameras               & 50.60 & 40.00 & 50.40 \\
			w/o consistency         & 51.40 & 41.00 & 51.00 \\
			\bottomrule
		\end{tabular}
		\label{tab:ablations}
	\end{table}

	
	\begin{table}[h!tp]
		\footnotesize
		\caption{Effect of the 2D Pose Estimation on Human3.6M.}
		\centering
		\begin{tabular}{ l | c c c }
			& MPJPE$\downarrow$     & PMPJPE$\downarrow$ & NMPJPE$\downarrow$ \\
			\hline
			Ground-truth 2D         & 43.0 & 33.1 & 41.1 \\
			Detectron           & 55.1 & 43.1 & 52.2 \\
			CPN \cite{chen2018cascaded}         & 50.6 & 40.0 & 48.8 \\
			
			\bottomrule
		\end{tabular}
		\label{tab:detector}
	\end{table}

	
	\begin{table}[h!tp]
		\footnotesize
		\caption{Effect of the Receptive Field on Human3.6M. }
		\centering
		\begin{tabular}{ l | c c c }
			& MPJPE$\downarrow$     & PMPJPE$\downarrow$ & NMPJPE$\downarrow$ \\
			\hline
			1 frame         & 54.7 & 43.4 & 52.6 \\
			27 frames          & 53.1 & 42.1 & 51.2 \\
			81 frames       & 51.2 & 41.0 & 49.4 \\
			243 frames       & 50.6 & 40.0 & 48.8 \\
			
			\bottomrule
		\end{tabular}
		\label{tab:receptive_field}
	\end{table}

	
	\begin{table*} [!htb]
		\footnotesize
		\caption{Evaluation results for protocol-II with the 243-frames temporal model and different levels of gaussian noise $\mathcal{N}(0,\sigma)$ ($\sigma$ is the standard deviation) added to the ground-truth 2D positions (\textit{GT}). All values are given in $mm$. }
		\centering
		\resizebox{0.99\textwidth}{!}{ 
			\begin{tabular}{l|ccccccccccccccc|c|ccc}
				\multicolumn{17}{c}{ } \\
				Protocol-II & Dir. & Dis. & Eat & Greet & Phone & Photo & Pose & Purch. & Sit & SitD & Smoke & Wait & WalkD & Walk & WalkT & Avg. \\
				\hline
				GT & 30.3 & 33.9 & 31.0 & 31.3 & 33.5 & 32.4 & 30.6 & 40.2 & 43.6 & 33.8 & 37.1 & 32.2 & 26.0 & 33.4 & 26.5 & 33.1   \\
				GT + $\mathcal{N}(0,5)$ & 30.4 & 34.1 & 30.7 & 33.0 & 34.8 & 32.5 & 31.2 & 40.7 & 47.7 & 34.8 & 39.0 & 31.8 & 26.8 & 35.6 & 27.5 & 34.1   \\
				GT + $\mathcal{N}(0,10)$ & 32.9 & 35.7 & 33.6 & 36.1 & 38.0 & 35.0 & 32.9 & 43.1 & 49.7 & 36.8 & 41.6 & 33.7 & 28.2 & 38.3 & 29.0 & 36.4   \\
				GT + $\mathcal{N}(0,15)$ & 35.4 & 37.9 & 36.5 & 39.1 & 39.7 & 37.5 & 35.6 & 45.6 & 53.6 & 39.2 & 44.9 & 36.8 & 29.8 & 40.8 & 31.4 & 39.0   \\
				GT + $\mathcal{N}(0,20)$ & 37.2 & 39.6 & 38.8 & 40.9 & 42.0 & 39.4 & 38.0 & 48.7 & 55.5 & 41.9 & 47.7 & 38.5 & 30.1 & 42.5 & 31.7 & 41.0  \\
				\bottomrule
		\end{tabular}}
		\label{tab:noise}
	\end{table*}

	
	\section{Conclusion}
	
	We presented an approach for 3D human pose estimation that explores the body pose temporal information combined with multi-view self-supervision. Our method requires a multi-view configuration only during training to obtain 3D body pose estimates by triangulation. The obtained pseudo-labels are then used to train a temporal convolutional neural network by additionally employing a geometric multi-view consistency constraint. During inference, our approach predicts the 3D body pose of a single individual from a sequence of 2D body pose estimates. In our experiments, we can achieve a performance that is competitive to fully-supervised learning. Without fine-tuning or retraining, our model is able to generalize to different scenes in the wild. Finally, we further examined the contribution of each loss function and the impact of different 2D body pose detectors.
	
	\section{ACKNOWLEDGMENTS}
	
	Part of this work was supported by the research project "KI Delta Learning" (project number: 19A19013A) funded by the Federal Ministry for Economic Affairs and Energy (BMWi) on the basis of a decision by the German Bundestag.
	

	{\small
		\bibliographystyle{ieee}
		\bibliography{egbib}
	}


\end{document}